\documentclass{article}

\usepackage{wrapfig}
\usepackage{multirow}
\usepackage{makecell}
\usepackage{array}
\usepackage{booktabs}
\usepackage{fancyvrb}
\usepackage{amsmath}
\usepackage{array}
\usepackage{longtable}
\usepackage{multirow}
\usepackage{fvextra}
\usepackage{tcolorbox}
\usepackage[boxed,ruled,vlined,linesnumbered]{algorithm2e}

\usepackage[utf8]{inputenc} 
\usepackage[T1]{fontenc}    
\usepackage{hyperref}       
\usepackage{url}            
\usepackage{booktabs}       
\usepackage{amsfonts}       
\usepackage{nicefrac}       
\usepackage{microtype}      
\usepackage{xcolor}         
\usepackage{graphicx}       

\usepackage{pifont}

\definecolor{redx}{RGB}{180,0,0}
\definecolor{greenx}{RGB}{0,180,0}

\newcommand{\redxmark}{\color{redx}\ding{55}}
\newcommand{\greencmark}{\color{greenx}\ding{51}}
\definecolor{redx}{RGB}{180,0,0}
\definecolor{greenx}{RGB}{0,180,0}

\usepackage{microtype}
\usepackage{graphicx}
\usepackage{subfigure}
\usepackage{booktabs} 

\usepackage{hyperref}

\usepackage{icml}

\newcommand{\goodname}{TorchOpera}

\icmltitlerunning{\goodname: A Compound AI System for LLM Safety}

\begin{document}

\twocolumn[
\icmltitle{\goodname: A Compound AI System for LLM Safety}




\begin{icmlauthorlist}
\icmlauthor{Shanshan Han, Zijian Hu, Alay Dilipbhai Shah, Han Jin, }{}
\icmlauthor{Yuhang Yao, Dimitris Stripelis, Zhaozhuo Xu, Chaoyang He}{}

\end{icmlauthorlist}


\icmlaffiliation{uci}{University of California, Irvine}
\icmlaffiliation{usc}{University of Southern California}
\icmlaffiliation{cmu}{Carnegie Mellon University}
\icmlaffiliation{stevens}{Stevens Institute of Technology }
\icmlaffiliation{to}{TensorOpera Inc.}


\icmlkeywords{Machine Learning, ICML}

\vskip 0.3in
]

\begin{abstract}
We introduce \goodname, a compound AI system for enhancing the safety and quality of prompts and responses for Large Language Models. \goodname~ensures that all user prompts are safe, contextually grounded, and effectively processed, while enhancing LLM responses to be relevant and high quality. 
\goodname~utilizes the vector database for contextual grounding, rule-based wrappers for flexible modifications, and specialized mechanisms for detecting and adjusting unsafe or incorrect content. We 
also provide a view of the compound AI system to reduce the computational cost. Extensive experiments show that \goodname~ensures the safety, reliability, and applicability of LLMs in real-world settings while maintaining the efficiency of LLM responses.

\end{abstract}

\section{Introduction}






Safeguarding the inference of Large Language Models (LLMs) is critical in real-world applications. Safety risks may exist in the user prompts as well as LLM responses during the users' interactions with LLMs. 
Malicious users may exploit the vulnerabilities of LLMs by interacting with them with crafted prompts, e.g., prompt injections~\cite{Liu2023PromptIA,Kumar2023CertifyingLS,Zhu2023PromptBenchTE,Chu2024ComprehensiveAO,Tedeschi2024ALERTAC,Zhao2024WeaktoStrongJO}. The outputs generated by LLMs can also be offensive, discriminatory, nonsensical, factually incorrect (e.g. hallucinations), etc~\cite{Zhang2023SirensSI,zhang2024efficient,Wang2023DecodingTrustAC,Fan2023OnTT,Huang2023ASO,Xu2024HallucinationII}, as the models are often trained on publicly available data that have uncontrolled sources and may include unsafe contents, which can propagate through the model's responses and produce harmful or inappropriate outputs.

Safeguarding LLMs is crucial and cannot be overstated. 
Unsafe inputs can not only manipulate the LLM outputs but also reveal sensitive information, bypass system instructions, or even execute harmful commands~\cite{Russinovich2024GreatNW,xu2024llm,chang2024play,Liu2023PromptIA,Kumar2023CertifyingLS,Zhu2023PromptBenchTE,Chu2024ComprehensiveAO,Tedeschi2024ALERTAC,Zhao2024WeaktoStrongJO}.
Also, problematic LLM outputs can confuse users, perpetuate negative stereotypes, and undermine public trust in LLM applications. 
This is particularly problematic in use cases where the accurate response is critical, such as healthcare, finance, and legal systems. 


Addressing the safety issues associated with LLM inference is complex, as safety risks can arise at any point when processing a user prompt and often necessitate coordination of different methods. 
To safeguard LLM inference from unsafe or problematic outputs, standalone approaches deploy ML models to detect different types of safety risks in the LLM outputs~\cite{Detoxify,perspective-api,openai-data-paper}. However, while these models can identify issues, they lack the functionality to correct errors directly within the outputs. This gap can be partially bridged by employing a separate model designed specifically for fixing detected errors. Nevertheless, this approach still fails to address certain challenges, such as hallucinations, which often stem from insufficient, inaccurate, or outdated source information~\cite{Xu2024HallucinationII,Zhang2023SirensSI,Huang2023ASO}.
In light of this, we can employ retrieval-augmented generation (RAG)~\cite{rag,chen2024benchmarking,gao2023retrieval}, which mitigates hallucinations by enriching the user query with additional contextual information from external data sources, thus effectively enhancing the LLM’s responses in more factual content.

\begin{table*}[ht]
\centering
\caption{Comparison of moderation-based harmfulness mitigation approaches}
\label{tab:moderation_comparison}
\scriptsize
\begin{tabular}{lccccccc@{}} 
\toprule
\textbf{Feature} & \textbf{Perspective API} & \textbf{Open AI} & \textbf{Nvidia NeMo} & \textbf{GuardRails} & \textbf{Detoxify} & \textbf{Llama Guard} & \textbf{Ours} \\ \midrule
Open-sourced &\redxmark&\redxmark&\greencmark &\greencmark &\greencmark &\greencmark &\greencmark \\
Self-developed model &\greencmark &\greencmark &\redxmark&\redxmark&\greencmark &\greencmark &\greencmark \\
Deployable on edge devices &\greencmark &\greencmark & - & - &\greencmark & \redxmark &\greencmark \\
Zero-shot generalization &\greencmark &\greencmark & - & - &\greencmark &\greencmark &\greencmark \\
Explainable results &\redxmark&\redxmark&\redxmark&\redxmark&\redxmark&\redxmark& \greencmark \\
Flexible workflow &\redxmark&\redxmark&\greencmark &\greencmark &\redxmark&\redxmark&\greencmark \\
\bottomrule
\end{tabular}
\end{table*}

However, all of these approaches fail to support quick customization of user needs and cannot handle rapidly changing contexts and system settings, while in certain cases, a simpler, code-based rule approach, denoted as ``wrappers'', can be more effective~\cite{guardrails}. 
As an example, consider a scenario where users want to add some texts at the beginning of the LLM outputs to warn if there are any phishing URLs in the texts. To achieve this goal, we can utilize crafted regex patterns to obtain URLs in the LLM outputs, and call APIs, such as Google Safe Browsing~\cite{google-safe-browsing} or PhishTank~\cite{phishtank}, to find out unsafe URLs. 
Such a method does not require model training or additional supplementary data sources and can be easily deployed in the LLM system, thus is more flexible to be quickly adapted to new safety challenges as they emerge.
Therefore, enhancing the overall safety of LLM inference requires a comprehensive, integrated system that orchestrates nodes with functionality components, such as ML models, RAG for grounding, and wrappers, while ensuring these components work collaboratively and harmoniously. 

This paper proposes \goodname, a compound AI system that enhances LLM safety throughout the entire lifecycle of LLM interactions. 
\goodname~systematically manages safety challenges in LLM applications comprehensively, thus 
advancing a safe and scalable deployment of LLM applications. It also adapts to new safety challenges with high flexibility, thus is suitable for changing user requirements and system environments.
\goodname~orchestrates several key nodes with different specializations, including 1) 
a Safety Detection Node that identifies safety risks in user inputs and LLM outputs, 2) a Grounding Node that utilizes vector databases to contextualize user queries, and 3) a Repair Node that leverages different methods, such as ML models and code-based wrappers, to correct errors detected in the LLM outputs. 

    

\textbf{Contribution. }We propose \goodname, a compound system to address safety issues in LLMs, instead of considering different safety challenges as sole and independent issues. \goodname~consists of multiple modules, including Safety Detector, Grounding, and Repairer, each targeting specific safety challenges in LLM interactions.
The core module, Safety Detector, is developed based on our own base model that has 1.6B parameters, thus is light-weighted and can be deployed on edge devices.
We distinguish between safety detection in user inputs and LLM outputs, specifically addressing hallucinations in LLM outputs. 
To facilitate further refinement of outputs that may contain hallucinations, we propose a model that not only detects hallucinations but also provides explanations based on the LLM's generative capabilities. These explanations can further be used to refine and improve the LLM responses.
TorchOpera allows user-defined wrappers to create flexible rules for fixing easily addressable errors in LLM-generated content, which enhances adaptability and customization of the system.

\section{Related Work}

Moderation-based harmfulness mitigation approaches ensure the outputs generated by LLMs are safe, appropriate, and free from harmful content~\cite{openai-data-paper,nemo,perspective-api,Detoxify,guardrails}. These methods leverage rule-based methods, machine learning classifiers, and human interfaces to monitor, evaluate, and manage the outputs produced by LLMs.

\textbf{Close-sourced solutions. }
OpenAI Moderation API~\cite{openai-data-paper} and Perspective API~\cite{perspective-api} utilize ML classifiers to detect undesired contents. These approaches provide scores for pre-defined categories of harmful content, such as toxicity, severe toxicity, identity attacks, insults, threats, etc. These tools are widely used in content moderation to filter out harmful content and has been incorporated into various online platforms to maintain safer user interactions~\cite{perspective-api-case-studies}. However, they are less adaptable to emerging safety risks as they are not open-sourced and cannot be finetuned.

\textbf{Opensourced solutions. }
LlamaGuard~\cite{inan2023llamaguard} leverages the zero/few-shot abilities of the Llama2-7B architecture~\cite{touvron2023llama} and can adapt to different taxonomies and sets of guidelines for different applications and users. Despite its adaptability, LlamaGuard's reliability depends on the LLM's understanding of the categories and the model's predictive accuracy. However, deploying LlamaGuard on edge devices is challenging due to its large number of parameters, which typically exceed the computing resources available on edge devices.
Detoxify~\cite{Detoxify} offers open-source models designed to detect toxic comments. These models, based on BERT~\cite{devlin2018bert} and RoBERTac~\cite{liu2019roberta} architectures, are trained on the Jigsaw datasets~\cite{jigsaw-unintended-bias-in-toxicity-classification,jigsaw-toxic-comment-classification,jigsaw-multilingual}. Detoxify provides pre-trained models that can be easily integrated into other systems to identify toxic content. Also, the models are able to recognize subtle nuances in language that might indicate harmful content, making them effective for moderating content .

\textbf{Customizable solutions. }
Guardrails~\cite{guardrails} and Nvidia NeMo Guardrails~\cite{nemo} employ customizable workflows to enhance LLM safety. 
Guardrails~\cite{guardrails} define flexible components, called ``rails'', to enable users to add wrappers at any stage of inference, which enables  users to add structure, type, and quality guarantees to LLMs outputs. Such rails can be code-based or using ML models. However, it does not have self-developed model and miss a unified solution for general cases. 
Nvidia NeMo Guardrails~\cite{nemo} functions as an intermediary layer that enhances the control and safety of LLMs. 
This framework includes pre-implemented moderation dedicated to fact-checking, hallucination prevention, and content moderation, which offers a robust solution for enhancing LLM safety.

We compare our approaches with the existing approaches in Table~\ref{tab:moderation_comparison}.

\begin{figure*}
  \centering
  \includegraphics[width=\textwidth]{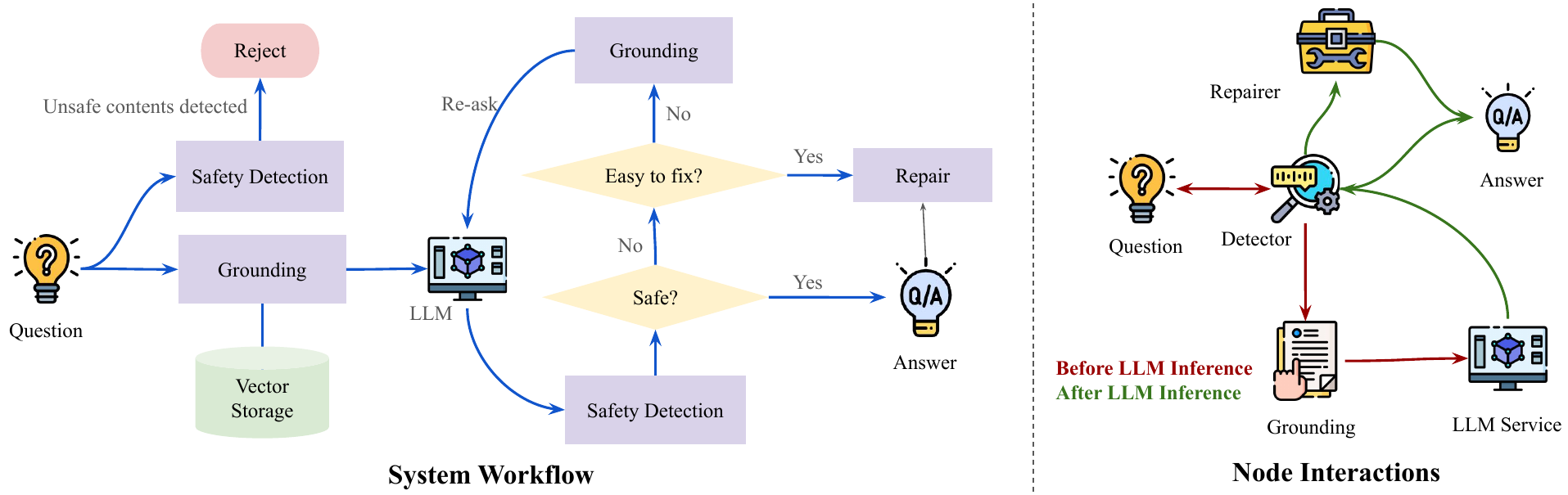}
  \caption{System Overview.}
  \label{fig: system_overview}
\end{figure*}
\section{\goodname~System Design}
\goodname~compound system enhances safety of LLM inputs and outputs while improving their quality. Specifically, it achieves two goals, 1) all user inputs are safe, contextually grounded, and effectively processed, such that the inputs to the LLMs are of high-quality and informative; and 2) the output generated by the LLMs are evaluated and enhanced, such that the outputs passed to users can be both relevant and of high quality. 
The workflow can be partitioned into two parts, including 
1) processing before LLM inference that enhances user queries, and 2) processing after LLM inference that detects undesired content and handle them properly. We overview our system in Figure~\ref{fig: system_overview}.


\textbf{1. Pre-inference processing. }
Before sending user queries to LLMs, \goodname~detects if there are any safety issues in the queries with \goodname~Safety Detector and ground the queries with context knowledge with \goodname~Grounding. 
\goodname~Safety Detector monitors user inputs to identify and reject queries that might be unsafe. The monitoring includes typical safety checks, including toxicity, stereotypes, threats, obscenities, prompt injection attacks, etc. Any form of unsafe content will lead to the queries being rejected. 
Inputs that pass this initial safety check are grounded with context with \goodname~Grounding, where the user query is tailored into an enhanced format by being contextualized with relevant knowledge retrieved from the vector data storage. By equipping the query with some context knowledge, the LLM can do inference with enriched information, thus can reduce hallucinations when generating responses. The details of \goodname~Safety Detector and \goodname~Grounding will be introduced in \S\ref{sec:safety_detector} and \S\ref{sec:grounding}, respectively.

\textbf{2. Post-inference processing.} 
Upon LLM finishing inference, \goodname~Safety Detector detects safety issues in the LLM outputs, specifically, hallucinations. This is because LLM applications typically leverages well-developed LLMs or APIs, such as LLaMA~\cite{touvron2023llama} and ChatGPT API~\cite{openai-data-paper}, which are generally safe and less likely to generate toxic or other unsafe content, while hallucinations occur frequently. The Safety Detector not only identifies hallucinations but also provides reasons for their occurrence, such that \goodname~can utilize the reasoning for later refinement of the LLM outputs. To achieve goal, \goodname~employs a text generation model to generate explainable results, and adjusts the loss function during training to ensure the model to produce classification results. 
\goodname Safety Detector also uses code to analyze the responses generated to derive classification results. 
After \goodname Safety Detector finishes detection, \goodname~Repair fixes the problematic content or aligns the outputs with some rule-based wrappers to meet user expectations. If the outputs are difficult to fix, such as those containing persistent hallucinations, 
the raw outputs are passed to \goodname~Grounding to generate a new prompt, which is then used to re-query the LLM. Details about \goodname~Repair can be found in \S\ref{sec:fixing}.

\section{\goodname~Components}
\subsection{\goodname~Safety Detector}\label{sec:safety_detector}


The Safety Detector addresses unsafe inputs and LLM outputs generated by LLMs to ensure both the ensuring that both the prompts provided to the models and the outputs they generate are free from harmful content and misinformation.


\textbf{Handling Unsafe Inputs. }
\goodname~utilizes a unified model specifically to detect undesired content in the user queries, such as toxicity, malicious prompts, stereotypes, threats, etc. One of the biggest challenges in training such model is the discrepancy between the training data and real-world user query distributions, where using traditional datasets alone can result in poor performance due to their divergence from actual user queries~\cite{openai-data-paper}.
To mitigate these issues, we crafted a training dataset by combining samples randomly selected from 15 public datasets. The datasets used include:
HEx-PHI~\cite{anonymous2024finetuning}, OpenAI~\cite{openai-data-paper}, 
Hotpot QA~\cite{yang2018hotpotqa},
Truthful QA~\cite{lin2021truthfulqa},
Awesome ChatGPT Prompts~\cite{awesome-chatgpt-prompts}, 
Jigsaw Unintended-Bias Data~\cite{jigsaw-unintended-bias-in-toxicity-classification},
GPT-Jailbreak~\cite{ChatGPT-Jailbreak-Prompts},
Jailbreak~\cite{jailbreak-classification},
Personalization Prompt~\cite{filtered_personalization_prompt_response},
QA Chat Prompts~\cite{qa-chat-prompts},
ChatGPT Prompts~\cite{ChatGPT-prompts},
10k Prompts Ranked~\cite{10k_prompts_ranked},
Iterative Prompt~\cite{iterative-prompt-v1-iter1-20K}, and
Instruction Following~\cite{instruction-following}.
By integrating data from multiple sources, including various domains and contexts, we can better simulate the variety of queries that users might submit.
Such sophisticated dataset capturing the diverse range of content types and user behaviors that LLMs are likely to encounter in practice, thus ensures a more representative sample of potential real-world inputs. 

\textbf{Handling Unsafe Outputs.} 
While current LLMs, such as ChatGPT~\cite{openai-data-paper} and Llama~\cite{touvron2023llama,inan2023llamaguard}, are generally designed to be safe and often refuse to answer queries that appear even slightly malicious, they can sometimes be overly cautious~\cite{openai-data-paper,inan2023llamaguard}. This conservative nature stems from the models' underlying safety protocols, which prioritize avoiding any potential harm over providing nuanced responses. Consequently, these LLMs may refuse to a wide range of queries that may not pose actual risks but fall within ambiguous or borderline categories. 
Given that many AI applications are currently built on top of well-established LLMs, their conservative nature reduces the necessity to detect toxicity in their outputs. 
Nevertheless, the most significant challenge remains the issue of hallucinations. While grounding can reduce the frequency of hallucinations by enriching the context available to the model, it cannot eliminate them entirely. This is because the model's inherent propensity to generate plausible-sounding but incorrect information is a fundamental characteristic that stems from its training on large, diverse datasets, which inevitably contain errors and inconsistencies.

To address hallucinations in LLM outputs, \goodname~Safety Detector utilizes a fine-tuned model designed to both detect hallucinations and provide hints for their subsequent mitigation. Specifically, the model is required to perform two key tasks:
1) classification: the model should detect the presence of hallucinations in LLM outputs; and 2) text generation: the model should generate explanations for why it identifies certain outputs as hallucinations to provide insights for subsequent output repair.

To achieve this goal, we employ the methods as follows. 

\begin{enumerate}
    \item Crafting Training Data. We use hallucination classification data and frame the questions and classification results as sentences. For example, the question might be, ``Is there any hallucination in the following record?'' with answers of ``Yes'' or ``No''. We also generate explanations of hallucinations with GPT4 API~\cite{openai-data-paper} and include them in the training data. 

    \item  Using Generative Models. We explore the generative abilities of LLMs. By fine-tuning generative models, we leverage their ability to generalize, allowing them to handle a variety of hallucination scenarios effectively.

    \item  Classification Results. \goodname~Safety Detector determines whether an output contains hallucinations based on ``Yes'' or ``No'' responses.
\end{enumerate}

\begin{figure*}[htpb]
    \centering
    \includegraphics[width=0.7\textwidth]{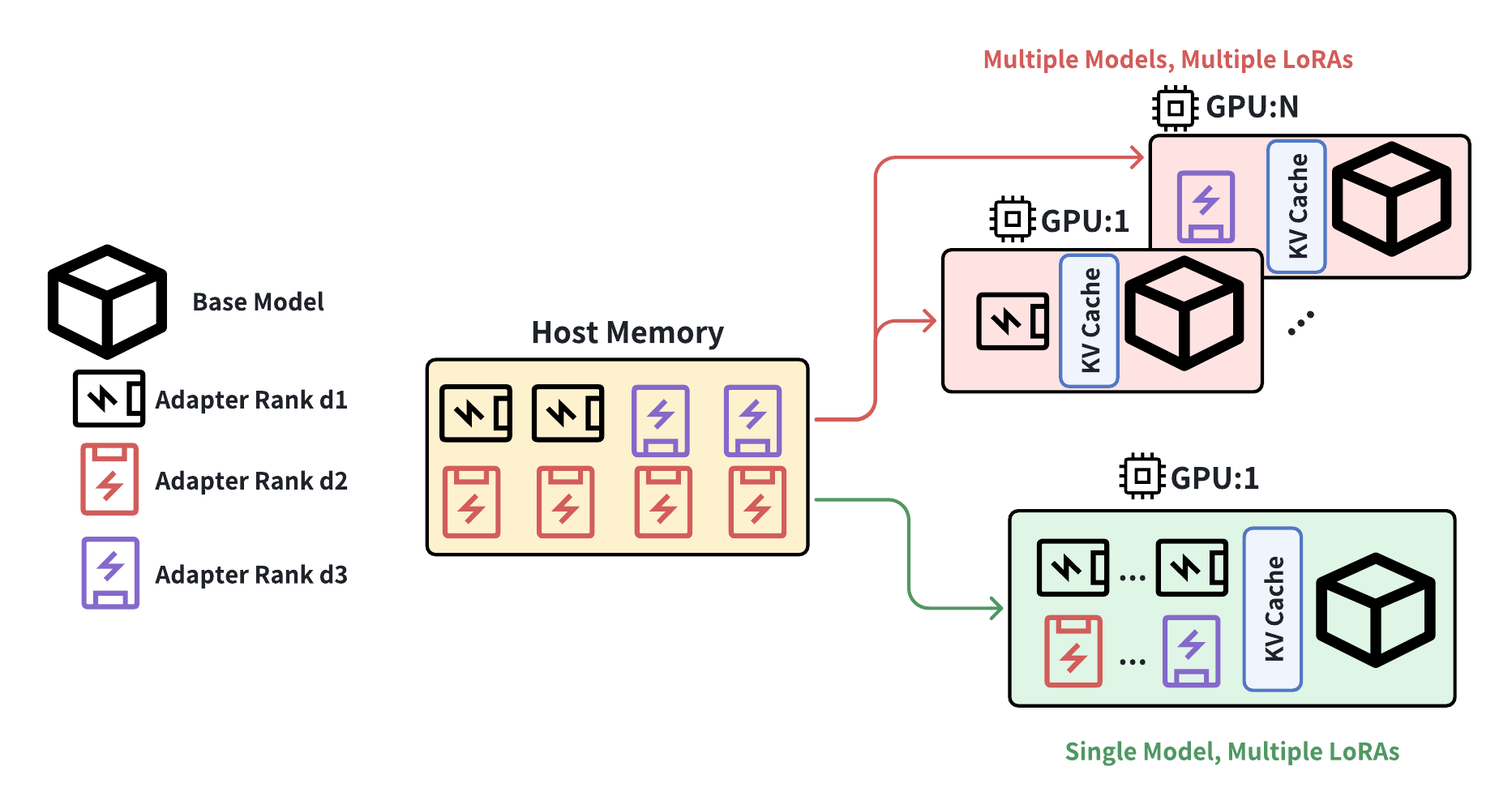}
    \caption{Serving multiple LoRAs with a single GPU.}
    \label{fig:multiple-loras}
\end{figure*}

\subsection{\goodname~Grounding}\label{sec:grounding}

\goodname~Grounding enhances the contextual richness and informativeness of user queries by leveraging external knowledge stored in vector data repositories. This process ensures that LLMs can utilize such knowledge to generate high-quality outputs, particularly by grounding user queries before they are passed to the LLMs for inference.

To support similarity search over the knowledge data, we create vector storage for plaintext knowledge. This involves vectorizing entire knowledge entries to create vector indexes. We employ two primary methods for indexing:
\begin{itemize}
    \item Whole Knowledge Indexing: This method involves indexing entire data from the datasets. While comprehensive, it reflects the data distribution and ensurers that the indexed data captures the contextual variety and complexity found in real-world queries.

\item Key Information Indexing: This method involves indexing only the questions. By focusing on key information, this approach targets the core elements of user queries, facilitating efficient retrieval of relevant data.
\end{itemize}

To ensure grounding to be effective, i.e., desired knowledge can be retrieved when issuing a query, the distribution of index should follow the distribution of real-world queries. This ensures that the model's responses are both relevant and practical in real-world applications. By capturing the distribution of actual user queries, we improve the model's ability to generate contextually appropriate responses. 




\begin{figure*}[htpb]
    \centering
    \includegraphics[width=0.96\textwidth]{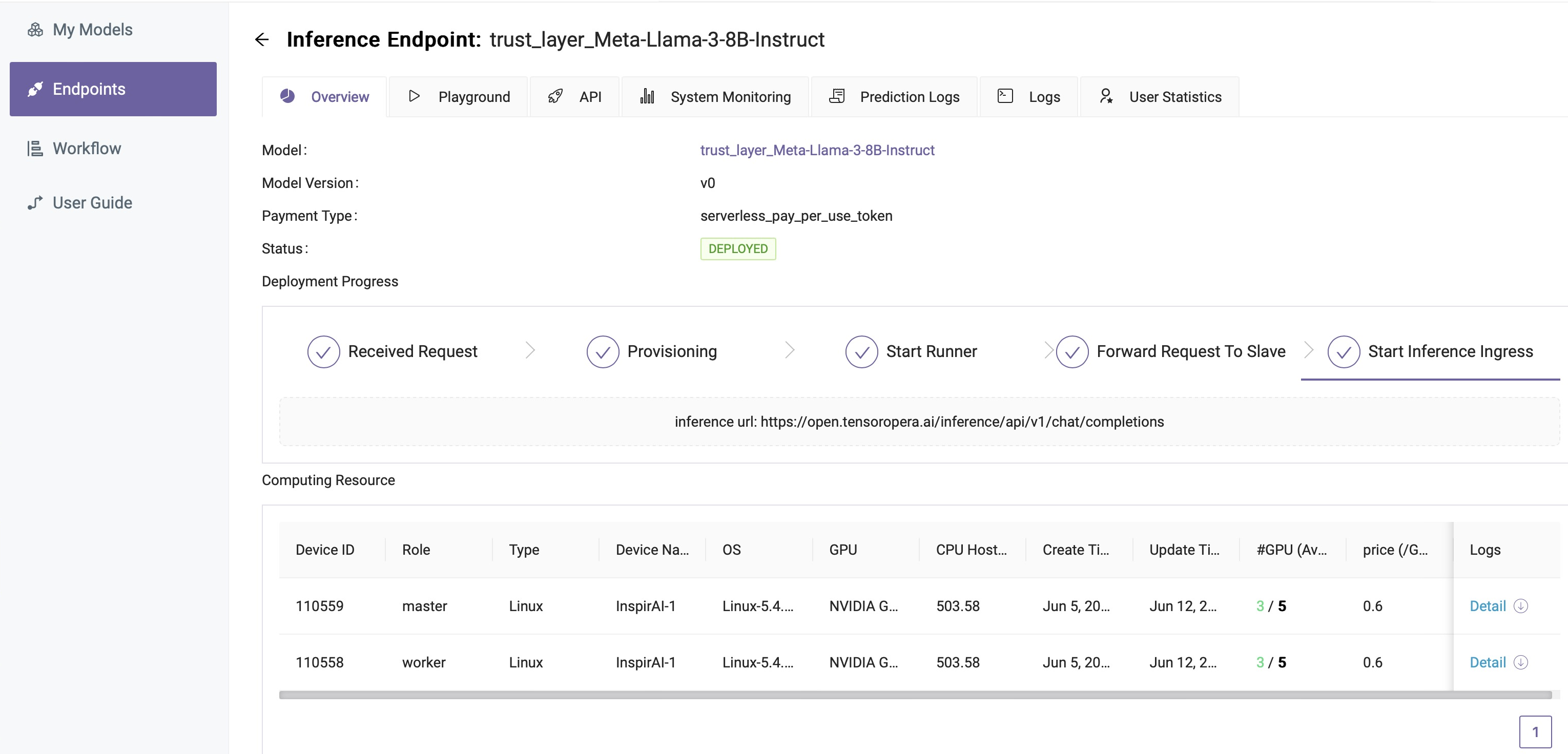}
    \caption{UI of defining \goodname~as a service in our platform.}
    \label{fig:ui}
\end{figure*}


\subsection{\goodname~Repair}\label{sec:fixing}
\goodname~Repair leverages light-weight wrappers to enable defining guidelines for fixing problematic contents in the LLM outputs with high flexibility. The wrapper can incoorporate code-based rules, APIs, web searching, and small models to accomplish editing and fixing tasks efficiently, which reduces computational overhead of calling costly LLMs, thus is more flexible, commercial, and easy to adapt.
Also, it simplifies and accelerates the development of user protocols, especially for scenarios that requires real-time adjustment to changing conditions where training or finetuning the inference LLMs is impractical, thereby enhances the adaptability to evolving user requirements and safety protocols and avoiding the intricacies involved with pretraning or finetuning LLMs.

\textbf{Warning URLs task example. }
The objective was to detect if LLM outputs contain URLs and prepend a warning message of the unsafe URLs at the beginning of the LLM outputs. \goodname~Repairer should check the safety of the URLs founded,  i.e., whether they are malicious or unreachable, and includes such information in the warning if they were unsafe.
Note that such task cannot be accomplished with prompt engineering when asking queries to LLMs, as the warning message should be at the beginning, and since LLMs generate contents token by token, the later contents in generation can not be predicted. To fix the LLM outputs with proper warning of URLs, \goodname~Repairer utilizes a regular expression pattern 
to identify URLs within the text. Upon URLs founded, \goodname~Repairer calls APIs for detecting phishing URLs, such as Google SafeBrowsing~\cite{google-safe-browsing}, and assess the accessibility of the benign URL by issuing web requests. Malicious URLs, as well as unreachable URLs that return status codes of 4XX, are added in the warning at the beginning of the LLM outputs.



\section{Implementation of~\goodname}
\subsection{Introduction of Base Model}\label{sec:base_model_intro}

Our base model is self-developed, decoder-only transformer-based small language model (SLM) with only 1.6B parameters~\footnote{We will provide more details after we release the base model.}. 
SLMs are gaining popularity in the AI landscape by offering powerful capabilities with minimal computational requirements. This evolution is particularly significant as it opens avenues for deploying AI in various settings, from mobile devices to server constraints, all while maintaining high performance. The integration of SLMs into innovative architectures such as mixture of experts (MoE) and model federation further amplifies their utility, enabling the construction of larger, more powerful systems by synergistically combining multiple (or many) SLMs. SLMs significantly reduce latency and require far less computational power compared to LLMs. This allows them to process queries more quickly, which enhances both the speed and cost-efficiency of inference, as well as responsiveness in generative AI applications.
Also, SLM can be deployed and trained across a myriad of platforms and devices, ranging from powerful GPUs on the cloud to edge devices like smartphones and AI PCs. This adaptability ensures that SLMs can be integrated into various technological environments, thus broadening their applicability and enhancing user accessibility.
Moreover, SLMs are particularly well-suited for integration into composite AI architectures such as Mixture of Experts (MoE) and model federation systems. These configurations utilize multiple SLMs in tandem to construct a more powerful model that can tackle more complex tasks like multilingual processing and predictive analytics from several data sources.

Our base model was trained with a 3-stage data curriculum on 3 trillion tokens of text and code data in 8K sequence length. The base model uses grouped query attention (GQA) with 4 KV heads and 16 attention heads and has a deeper architecture than other SLMs. Specifically, our base model has 32 transformer decoder blocks which is 78\% deeper than Gemma-2B~\cite{team2024gemma}, 33\% deeper than Qwen1.5-1.8B~\cite{qwen} and StableLM-2-1.6B~\cite{bellagente2024stable}, and 15\% deeper than OpenELM-1.1B~\cite{mehta2022cvnets,mehtaOpenELMEfficientLanguage2024}.

\subsection{Efficient Deployment of Safety Detectors}

Compared to traditional approaches that serve LoRA-based fine tuned models by copying the same pre-trained model for every other LoRA adapters across multiple GPUs, we reuse the same base model copy to serve multiple LoRAs using a single GPU. Figure~\ref{fig:multiple-loras} provides an architectural overview of \goodname~'s single model, multi-LoRA serving system, as well as a comparison to the multi-model, multi-LoRA counterpart. In our implementation, we build on top of the Punica~\cite{chen2024punica} multi-tenant LoRA serving system that supports continuous batching for adapters of heterogeneous ranks and enables the dynamic and on-demand loading of multiple heterogeneous adapters to meet the demand of various request workloads.

This deployment not only leads to enhanced and efficient utilization of the available GPU resources but also reduces the computational costs. 
By minimizing the redundancy of model replication, \goodname also achieves higher scalability and flexibility, making it possible to extend our system to applications using multiple adapters in the future. Additionally, such dynamic adapter loading mechanism ensures optimal response time and efficient resource allocation, which can serve multiple models of the compound system in demanding computational environments.

\begin{figure*}
  \centering
  \includegraphics[width=\textwidth]{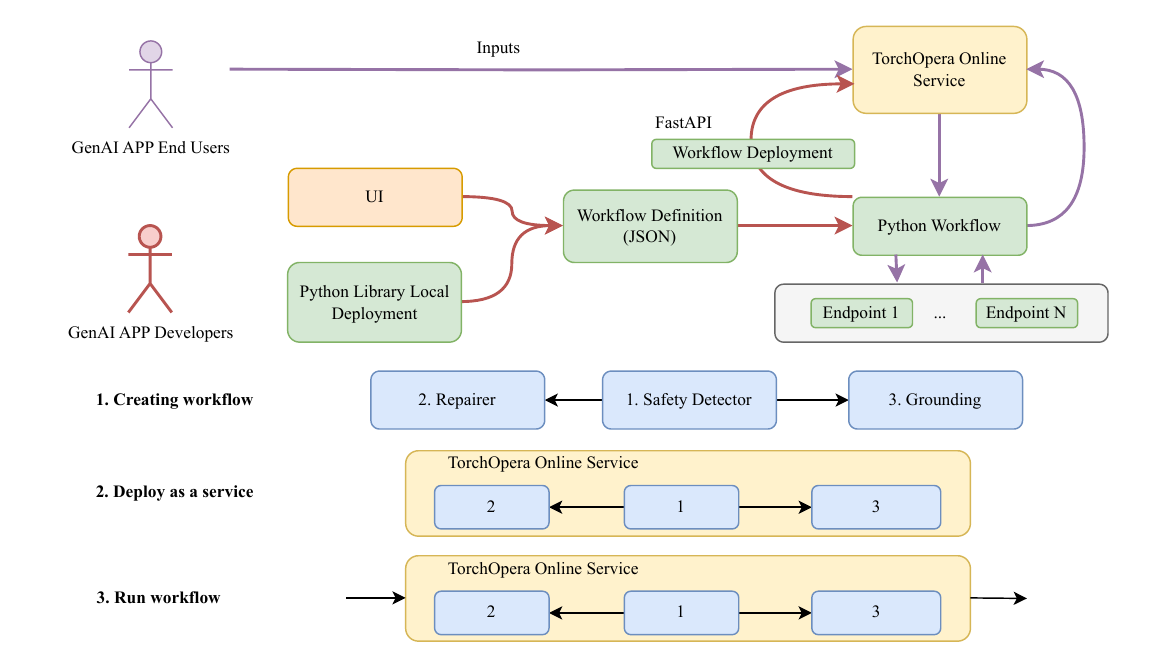}
  \caption{\goodname~coumpound system as a service. \textcolor{red}{Red lines} indicate deployment flow. \textcolor{purple}{Purple lines} indicate service flow.}
  \label{fig: torchopera_service}
\end{figure*}

\section{\goodname~Compound System as a Service}\label{sec:service}
We have a platform that allows users to define, deploy and serve compound AI system based applications efficiently. Figure~\ref{fig:ui} shows the UI of deploying \goodname to safeguard an inference model Llama-3-8B-instruct~\cite{touvron2023llama}. 
From a system perspective, we facilitate cross-cloud execution across a decentralized network of nodes with a scalable and cloud-agnostic compute layer. This enables  the management of compute resources efficiently,  without geographic constraints. 
Also, our infrastructure can dynamically adjust the compute resources of each endpoint, based on real-time workload demands. This ensures  responsiveness and availability of each endpoint, making our infrastructure to handle varying workload efficiently while reducing unnecessary computational costs.

\begin{figure}
  \centering
    \includegraphics[width=0.45\textwidth]{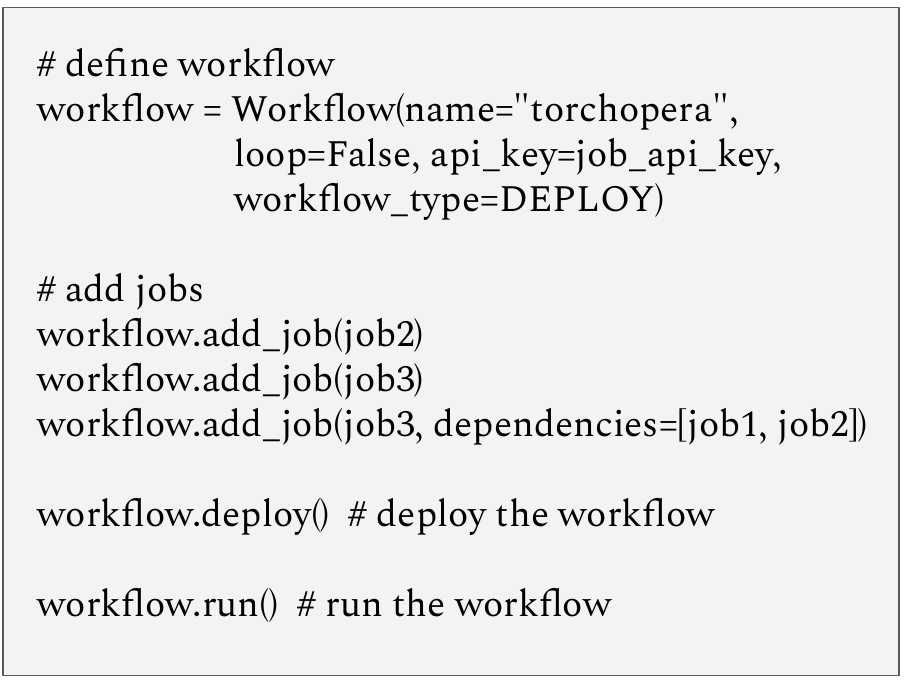}
    \caption{Key steps of deploying \goodname.}
    \label{fig:deploy_code}
\end{figure}

Figure~\ref{fig: torchopera_service} presents the workflow of deploying \goodname~as a service and calling \goodname~for safeguarding LLM inferences. To deploy \goodname, developers define components of \goodname, i.e., Safety Detector, Grounding, and Repairer, then deploy \goodname, and run the workflow to serve it. 
We provide key steps of deploying \goodname~in Figure~\ref{fig:deploy_code}.
To call \goodname~service, GenAI APP users (or some APP inputs) trigger the service, which will call necessary counterparts for inferences with Python backends; see purple lines in Figure~\ref{fig: torchopera_service}.

\section{Conclusion}\label{sec:conclusion}
In this paper, we introduced \goodname, a comprehensive AI system designed to enhance the safety and reliability of large language models (LLMs). By integrating several key components, including Safety Detector, Grounding, and Repairer, we addressed critical challenges such as hallucinations, unsafe inputs, and the need for contextually enriched outputs.
Our Safety Detector plays a pivotal role in ensuring user safety by detecting various forms of harmful content in user inputs and outputs. We demonstrated the importance of crafting training datasets that reflect real-world query distributions to improve the model's accuracy and reliability. We also propse a finetuned model that achieves dual functionalities of both classification and text generation, which allows for not only the identification of hallucinations but also the provision of detailed explanations, which provides hints for further repair of hallucinations. We believe that our work can serve as a stepping stone to contribute  to the advancement of the field of LLM safety.

\bibliography{main}
\bibliographystyle{icml}

\end{document}